\documentclass[letterpaper, 10 pt, conference]{ieeeconf}  

\IEEEoverridecommandlockouts                              

\makeatletter
\let\NAT@parse\undefined
\makeatother
\usepackage[numbers,sort&compress]{natbib}

\usepackage{amsmath, bm} 
\usepackage{amssymb}
\DeclareMathOperator*{\argmax}{arg\,max}

\let\labelindent\relax
\usepackage{enumitem}
\usepackage{graphicx}
\usepackage[ruled,vlined,linesnumbered]{algorithm2e}
\usepackage{setspace}
\usepackage{xcolor}
\usepackage{xurl}

\title{\LARGE \bf
Interpreting Contact Interactions to Overcome Failure\\ in Robot Assembly Tasks
}
\author{Peter A. Zachares$^{1}$, Michelle A. Lee$^{1}$, Wenzhao Lian$^{2}$, Jeannette Bohg$^{1}$
\thanks{1. Department of Computer Science, Stanford University, 2. X, The Moonshot Factory}
\thanks{This work has been partially supported by JD.com American Technologies Corporation (“JD”) under the SAIL-JD AI Research Initiative, by the Toyota Research Institute ("TRI") and by X, The Moonshot Factory. This article solely reflects the opinions and conclusions of its authors and not of JD, any entity associated with JD.com, X, The Moonshot Factory, TRI, or any entity associated with Toyota.}%
}
\begin{document}
\maketitle
\begin{abstract}
A key challenge towards autonomous multi-part object assembly is robust sensorimotor control under uncertainty. In contrast to previous works that rely on \textit{a priori} knowledge on whether two parts match, we aim to learn this through physical interaction. We propose a hierarchical approach that enables a robot to autonomously assemble parts while being uncertain about part types and positions. In particular, our probabilistic approach learns a set of differentiable filters that leverage the tactile sensorimotor trace from failed assembly attempts to update its belief about part position and type. This enables a robot to overcome assembly failure. We demonstrate the effectiveness of our approach on a set of object fitting tasks. The experimental results show that the proposed approach achieves higher precision in object position and type estimation, and accomplishes object fitting tasks faster than baselines.
\end{abstract}

\section{Introduction}
Automobiles, computers, and even ballpoint pens are composed of multi-part assemblies~\cite{kimble2020benchmarking}. These products are relatively complex to manufacture and often require a team of humans and robots to complete a sequence of contact-rich manipulation tasks. While it would be beneficial if a robot could autonomously perform these multi-part assemblies, the required complex manipulation tasks are typically only successful in highly-controlled settings with little uncertainty~\cite{uncertainty0, uncertainty1, uncertainty2, Bohg2012a}. Consequently, one of the main challenges of multi-part assembly is robust sensorimotor control under uncertainty, which can be caused by inaccuracies in a robot's perception or control system and manufacturing flaws of the parts. 
An example real-world task is to fit an unlabelled set of screws into holes where their diameters vary on a sub-millimeter scale - a scale beyond the resolution of most affordable camera systems. In this scenario, the robot may easily fail to fit together two objects, but not know whether it failed because the objects do not match, \emph{or} because its manipulation strategy was ineffective. Understanding the reason for failure is crucial for eventual task success. 

In this paper, we propose an approach that enables a robot to  autonomously  assemble  objects  while  being  uncertain about object types and positions. Our insight is that a robot can learn something about an object when physically interacting  with  it  during  assembly.  Specifically,  sensorimotor traces  from  touch  sensing  can  give  additional  and  more precise information about the physical properties of objects than vision alone~\cite{neural_synergy}. Even if the robot does not succeed in fitting two objects together, it can then update its belief about the object type and position. This updated belief will inform decision-making for choosing the next assembly action. 

\begin{figure}
    \centering
    \includegraphics[width=\linewidth]{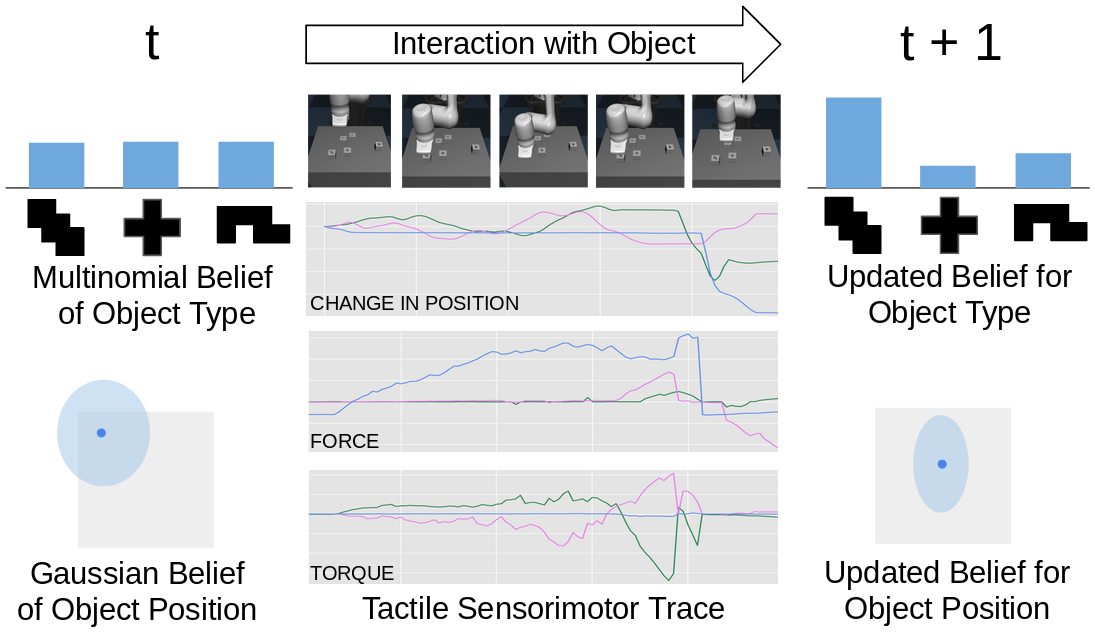}
    \caption{To fit objects together, our method maintains a belief about the positions and types of objects in the environment, and uses tactile sensorimotor traces from (failed) object interactions to update the belief.
    }
    \label{fig:teaser}
\end{figure}

We formulate the problem of multi-object assembly in a hierarchical way. The robot maintains a belief over the type and position of each object in the environment. The belief is initialized from vision. Based on this belief, the high level policy decides which object is most likely matching and attempts to fit the grasped object with the selected object using a low level policy. If the attempt fails, the robot updates its belief based on the collected sensorimotor traces (see Fig.~\ref{fig:teaser}).We designed a network architecture that takes sensorimotor traces as input and outputs an estimate of object type and position.
This information is used within differentiable filters~\cite{backprop_kf, histogram} to update the belief.
Given the updated belief, the robot can reason about the cause of failure. 
If the object was not matching, then the robot should attempt to fit a different object. If the object position estimate was poor but the object was a match, the robot should make another attempt to fit with the same object.

\textcolor{black}{We evaluate our approach on the assembly task instance of} a peg-in-hole problem with multiple pegs and holes of different shapes. A peg \& hole pair can only fit if their shapes match. The high level policy is used for state estimation and hole selection. Experimental results demonstrate that when using the proposed differentiable filters, fewer fitting attempts are needed to accomplish the full assembly task than with baseline methods. 
\section{Related Work}
\subsection{Object Fitting Tasks}
Object fitting tasks such as peg insertion, fastening screws and dense box packing have been studied for decades due to their relevance in logistics and manufacturing \cite{1988_peginsertion,cad_plus_rl_peginsertion,CADmodel_based_peginsertion,discrete_peginsertion,guapo_peginsertion,lee2019making,meta_peginsertion,mp_vision_fastrl_peginsertion,residual_peginsertion,rl_highprecisionpeginsertion,rl_socketinsertion,rl_softpeginsertion,zakka2019form2fit_peginsertion, spiral_search,tactileboxpacking}. Robust sensorimotor control under uncertainty has been identified as a major challenge early on. For instance, \citet{1988_peginsertion} proposed to use compliance and force-sensing to compensate for uncertainty due to noisy pose estimation, inaccurate control or manufacturing flaws. While \citep{1988_peginsertion} recognizes the importance of haptic feedback, it does not consider multi-part assemblies and the case where the peg and hole do not match.
More recently, works that leverage deep reinforcement learning (RL) learn manipulation policies for peg insertion directly from raw sensory data~\cite{discrete_peginsertion, lee2019making, mp_vision_fastrl_peginsertion, rl_highprecisionpeginsertion, rl_socketinsertion}. However, RL policies are often sample inefficient and take many interactions with the environment just to learn a single object fitting policy. Thus, it is not straightforward to adopt these methods for multi-part assemblies and the case  where the robot is uncertain whether two objects match. Consequently, we seek to develop a probabilistic approach that is not as data-hungry as RL methods. \citet{cad_plus_rl_peginsertion} propose a more sample-efficient RL method that relies on object CAD models and motion planning. However, the authors also do not consider the case where an object part may be misdetected and the robot has to recover from the resulting manipulation failure.
There is some prior work on recovering from failure in object fitting tasks. \citet{tactileboxpacking} use a high-resolution tactile sensor to perform dense box packing. The robot can reattempt the task and estimates the position of gaps in a densely packed box using tactile feedback. However, their approach assumes that the grasped object can eventually fit in the box, i.e. there is no uncertainty on object matching. \citet{zakka2019form2fit_peginsertion} approached the problem of {\em kitting\/} which is a multi-object fitting task. Here, the robot does not know with certainty the types of objects in the environment or how they fit. The authors learned an assembly policy for this task. Unlike our approach, their policy can only recover from failure through human intervention.
\subsection{Understanding and Learning From Failure}


%




There are two classes of techniques within the RL and decision-making literature that are most relevant to the problem we are considering: estimating an unknown reward function, and estimating the partially observed state online. Each of them allows to learn online from failure.

In multi-armed bandit (MAB) \cite{mab} and contextual multi-armed bandit \cite{contextualmab} problems, an agent learns the reward function for each action online; trying to minimize the regret created by its decision making. For example, \citet{mab_regrasp,dexnet1.0} formulated the problem of learning to grasp as a multi-armed bandit. Therefore, a robot can improve its ability to grasp an object after each attempt. \textcolor{black}{However, a MAB formulation applied to our task would necessitate using a single univariate distribution to describe uncertainty about the type and position of an object; making it difficult to disambiguate the cause of a failure.}


In Bayes-adaptive Markov decision processes (BAMDPs)~\cite{bayesadaptive_mdp} and partially observable Markov decision processes (POMDPs) \cite{pomdp_original}, the state of the environment is partially observed and estimated online. This framework has been used in robotic problems to learn online from failed execution~\cite{mab_regrasp, dexnet1.0, regrasp_POMDP, gilwoo}. 
For instance, \citet{regrasp_POMDP} used a POMDP formulation to improve a robot's ability to grasp an object online. \citet{gilwoo} combined a BAMDP problem formulation with a policy gradient method to learn a policy which also updates its belief about the state of the environment  online. 
Like our approach, \citet{gilwoo} learned a differentiable filter to estimate the partially observed environment state. 
\textcolor{black}{However for our approach, the differentiable filters are trained separately from the agent’s policy; making them agnostic to the agent's choice of policy}

\section{Problem Formulation}
\label{sec:problem}

We approach the problem of object fitting: the robot must find one out of $n$ objects in the environment that matches its grasped object and fit the pair. Uncertainty can cause task failure on both levels of the task: for the high level task, the robot does not know exactly the types and positions of the $n$ objects; for the low level task, the robot cannot reliably succeed in fitting the object pair even if they match. In the remainder of the paper, we use peg and hole to refer to the grasped object and an object in the environment, respectively. \textcolor{black}{However, this approach generalizes to any multi-object assembly task that needs to take into account object fit.}

For the high level task, we assume a noisy vision sensor that detects the types $c^i$ (e.g. shape or dimension) and positions $\mathbf{p}^i$ of each hole where $i=1, \cdots, n$. $c^i\in \{1, \cdots, C\}$, where $C$ is the number of hole types, and $\mathbf{p}^i \in \mathbb{R}^{2}$ as the holes are assumed to vary their positions within a known 2D plane. 
During assembly, the holes remain static. Thus, given a peg, the high level task is to find a matching hole, i.e., its index out of $\{1, \cdots, n\}$ and its position $\mathbf{p}^{match}$. We formulate this as a finite horizon partially observable Markov decision process (POMDP) with states $s \in \mathcal{S}$, observations $o \in \mathcal{O}$, actions $a \in \mathcal{A}$, horizon $K$, and reward function $r: \mathcal{S} \times \mathcal{A} \rightarrow \mathbb{R}$. The state space describes three properties of each of the $n$ holes in the environment: (i) the hole type $c$, (ii) the hole position $\mathbf{p}$, and (iii) a binary indicator $\beta$ of whether the peg is correctly fitted with this hole or not.
The action $a$ corresponds to choosing one of the $n$ holes in the environment\footnote{The low level policy requires both the type and position of a hole as input, but we assume its effectiveness is independent of the hole position given the hole type, thus simplifying our POMDP formulation.} i.e. $a \in \{1, \cdots, n\}$. The agent receives a reward if it correctly fits its peg into a matching hole. 

For solving this POMDP, we have to devise a policy for action selection that maximizes the reward based on the agent's current belief about the state of the environment $b(s)$, i.e., finding  $ \pi^{*} = \argmax_{\pi} \sum_{t=0}^{K} E\big[r(s_t, a_t)\big]$ to correctly fit the peg with a matching hole.

Once a hole is selected by the high level policy, we run a low level policy $\pi_{low}$ to attempt fitting. $\pi_{low}$ controls the robot's end-effector position $\mathbf{x}$ with control inputs $\mathbf{u}_{1:J}$ where $\mathbf{u}$ is a desired change in end-effector position and $J$ is the fixed horizon of the low level task. 
During policy execution, the robot collects a tactile sensorimotor trace $\mathbf{z}_{1:J}$ which is composed of the measurements from a wrist-mounted Force/Torque sensor, the end-effector position and orientation, the end-effector linear and angular velocity, and the binary contact state of the end-effector with the environment for each time step. Our hypothesis is that this trace provides information about the hole type and position that the high level policy can use to update its belief state.

\section{Method}

As described in Section \ref{sec:problem}, the state $s$ of each hole is not fully observable, so the robot maintains a belief $b_t^i(s)$, where $t$ describes the time step of the high level task and $i$ refers to the index of holes. We use a Gaussian belief $\mathbf{p}^i \sim \mathcal{N}(\bm{\mu}_t^i, \mathbf{\Sigma}_t^i)$ to model a hole position and a multinomial belief $c^i \sim Multi(\bm{\xi}_t^i)$ for a hole type, where $\sum_{c=1}^C \xi_{t,c}^i=1$. The initial belief is based on observations from a vision detector, which outputs noisy estimates of hole positions and types. The rest of the state space is fully observed, including the binary indicator $\beta^i$ of whether fit has been achieved with hole $i$, and $c^{peg}$, the type of the peg.

The reward function solely depends on the fully observed $\beta$ for all holes, and the action $a$ (the selected hole), as below:
\begin{equation}
r(s_t, a) = 
\begin{cases}
1\text{,} & \text{if } \beta_{t}^a = 0 \land \beta_{t+1}^a = 1;\\
0\text{,}  & \text{otherwise.}
\end{cases}
\end{equation}
This means the robot receives a reward of $1$ if it achieves fit with its current hole choice and $0$ otherwise. We assume the low level policy $\pi_{low}$ has a nonzero probability $\alpha$ of successfully fitting the robot's peg with the chosen hole if the peg and hole match. After choosing a hole $a$ the robot performs a rollout of $\pi_{low}$, and $\beta^a$ is updated depending on the rollout outcome. Thus, the dynamics model for the high level task is described as:
\begin{equation} \label{eq:dynamics_model}
T(s_{t+1}^a |s_{t}^a, c^{peg}) =
\begin{cases}
\alpha \mathbb{I}_{c^a=c^{peg}}, \\
\qquad\text{if } \beta_{t+1}^a = 1 \land \beta_{t}^a = 0 \text{;}\\ 
1 - \alpha\mathbb{I}_{c^a=c^{peg}}, \\
\qquad \text{if } \beta_{t+1}^a = 0 \land \beta_{t}^a = 0,\\
\end{cases}
\end{equation}
where $\mathbb{I}_{c^a=c^{peg}}$ is an indicator function of whether the type of hole $a$ matches the robot's peg type.

\subsection{High Level Task Policy}
\label{subsection:task}
\begin{figure}
    \centering
    \includegraphics[width=\columnwidth]{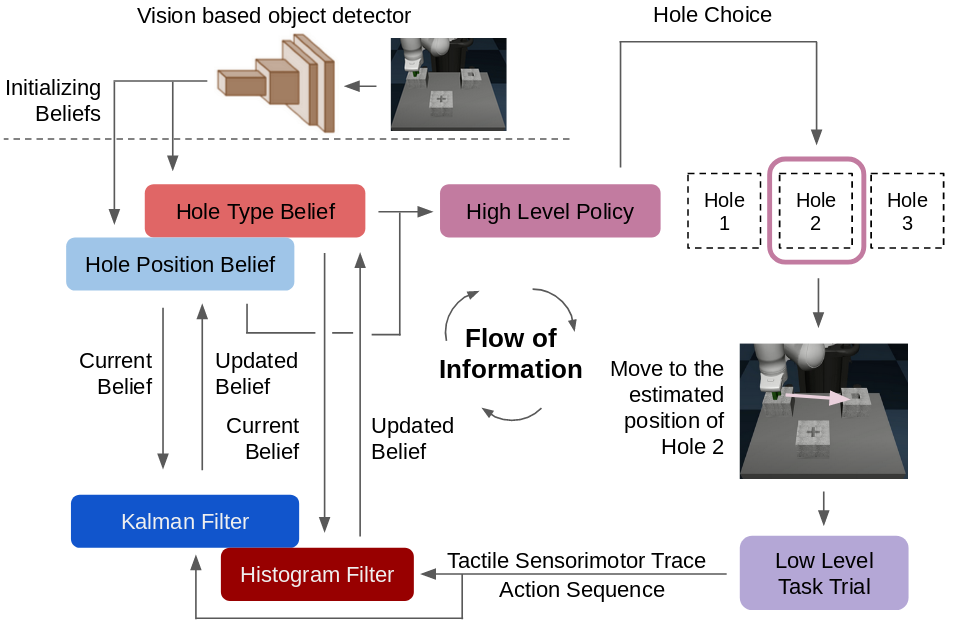}
    \caption{Overview of Approach. Starting with a vision prior, the robot chooses an object based on its current belief. If the robot fails to achieve fit, it uses tactile feedback from the interaction to update its belief about the environment and then decides which object to choose next.}
    \label{fig:system}
\end{figure}

In our problem, the hole states are independent of each other. Therefore, the robot cannot gain any additional information when interacting with holes which do not match the peg. Consequently, the robot can maximize its expected returns by greedily choosing the hole with which it has the highest probability of achieving fit based on its current belief:
\begin{equation}
    \pi^* = \argmax_{a \in \{1,\cdots,n\}} p(\beta_{t+1}^a = 1| b_{t}^a(s_t), c^{peg}),
\end{equation}
where $p(\beta_{t+1}^a = 1| b_{t}^a(s_t), c^{peg})=\sum_{c \in C} \alpha\mathbb{I}_{c = c^{peg}}  \xi_{t,c}^a$.
The overall high level policy is illustrated in Fig. \ref{fig:system}. As outlined in Algorithm 1,
our proposed high level policy performs the following steps at each iteration:
\begin{enumerate}[wide, labelindent=0pt]
    \item Choose a hole $i$ from $n$ holes (Line $3$).
    \item Move to the estimated hole position $\bm{\mu}_{t}^i$ (Line 4).
    \item Roll-out the low level policy $\pi_{low}$ (Lines 5-8).
    \item Update belief about hole position and type (Lines 9-11).
\end{enumerate}

To update the hole position estimates, the robot uses a differentiable Kalman filter \cite{backprop_kf}. Let $\mathbf{o}_{pos}^i$ denote a noisy observation of the true object position $\mathbf{p}^{i}$. We model observation noise as zero-mean Gaussian with covariance matrix $\mathbf{R}$ that we
estimate during training. 
We propose a virtual sensor $h_{pos}()$ that takes
the tactile sensorimotor trace as input and provides an unbiased estimate of the error between the current position estimate $\bm{\mu}_{t}^i$ and the true position $\mathbf{p}^{i}$, i.e. the innovation in a Kalman update:
\begin{equation}
h_{pos}(\mathbf{z}_{1:J}, \mathbf{u}_{1:J}, c^{peg}, \bm{\mu}_{t}^i) = \mathbf{o}_{pos}^i-\bm{\mu}_{t}^i.
\end{equation}

We assume fixed hole positions in our task, so the dynamics of the environment with respect to estimating position is an identity mapping. The resulting Kalman update for hole position belief is summarized in Algorithm 2.



\begin{algorithm}[t]
\setstretch{1.01}
\SetAlgoLined
\LinesNumbered
\KwData{$c^{peg}, b_{0}^{1:n}$}
 $t = 0$;\\
 \While{$\sum_{m=1}^n \beta_{t}^{m} = 0$}{
  $i = \argmax_{a \in \{1,\cdots,n\}} \hspace{1mm} \sum_{c \in C} \alpha\mathbb{I}_{c = c^{peg}}  \xi_{t,c}^a$;
  
  move to $\bm{\mu}_{t}^{i}$;
  
  \For{$ j \gets 0$ \KwTo $J$}{
    perform action $\mathbf{u}_j \gets \pi_{low}$;
    }
    
  observe $\mathbf{z}_{1:J}$, $\mathbf{u}_{1:J}$, $\beta_{t+1}^i$;
  
  $\bm{\mu}_{t+1}^i, \mathbf{\Sigma}_{t+1}^i$ = $update_{pos}$($\bm{\mu}_{t}^i, \mathbf{\Sigma}_{t}^i, \mathbf{z}_{1:J}, \mathbf{u}_{1:J}, c^{peg}$);
  
  $\bm{\xi}_{t+1}^i$ = $update_{type}$($\bm{\xi}_{t}^i, \mathbf{z}_{1:J}, \mathbf{u}_{1:J}, c^{peg}$, $\beta_{t+1}^i$);
  
  $t = t + 1$;}
\caption{High Level Policy}
\end{algorithm}
\begin{algorithm}[t]
\setstretch{1.1}
\SetAlgoLined
\LinesNumbered
\KwResult{$\bm{\mu}_{t+1}^i, \mathbf{\Sigma}_{t+1}^i$}
\KwData{$\bm{\mu}_{t}^i, \mathbf{\Sigma}_{t}^i, \mathbf{z}_{1:J}, \mathbf{u}_{1:J}, c^{peg}$}
$\mathbf{K} = \mathbf{\Sigma}_{t}^i (\mathbf{R} + \mathbf{\Sigma}_{t}^i)^{-1}$;\\

$\bm{\mu}_{t+1}^i = \bm{\mu}_{t}^i + \mathbf{K} h_{pos}(\mathbf{z}_{1:J}, \mathbf{u}_{1:J}, c^{peg}, \bm{\mu}_{t}^i)$;\\
$\mathbf{\Sigma}_{t+1}^i = (\mathbf{I} - \mathbf{K})\mathbf{\Sigma}_{t}^i$;\\
\caption{\hspace{-0.5mm}Object Position Estimation $update_{pos}()$}
\end{algorithm}

To estimate the type of a hole, the robot uses a differentiable histogram filter \cite{histogram} with a learned virtual sensor $h_{match}()$ and observation model $H(o_{match}|c^i, c^{peg})$. The learned virtual sensor $h_{match}()$ produces an estimate $o_{match}$ about whether the robot's peg matches with the hole it is interacting with. The observation model $H(o_{match}|c^i, c^{peg})$ describes the confusion matrix that we estimated during the training stage, which approximates the observation noise. Algorithm 3 summarizes the resulting update of the robot's belief about the type of a hole $i$ after a failed attempt.

\begin{algorithm}[t]
\setstretch{1.1}
\SetAlgoLined
\LinesNumbered
\KwResult{$\bm{\xi}_{t+1}^i$}
\KwData{$\bm{\xi}_{t}^i, \mathbf{z}_{1:J}, \mathbf{u}_{1:J}, c^{peg}$, $\beta_{t+1}^i$}
$o_{match} = h_{match}(\mathbf{z}_{1:J}, \mathbf{u}_{1:J}, c^{peg})$;\\
$\eta = \sum_{c \in \mathcal{C}} H(o_{match}|c, c^{peg})T(\beta_{t+1}^i|c,c^{peg})\xi_{t,c}^i$;\\
$\xi_{t+1,c}^i = \eta^{-1} H(o_{match}|c, c^{peg})T(\beta_{t+1}^i|c,c^{peg})\xi_{t,c}^i$;\\
\caption{Object Type Estimation $update_{type}()$}   
\end{algorithm}

\subsection{Model Architecture and Learning}
\label{section:model_learning}

\begin{figure}
    \centering
    \includegraphics[width=\columnwidth]{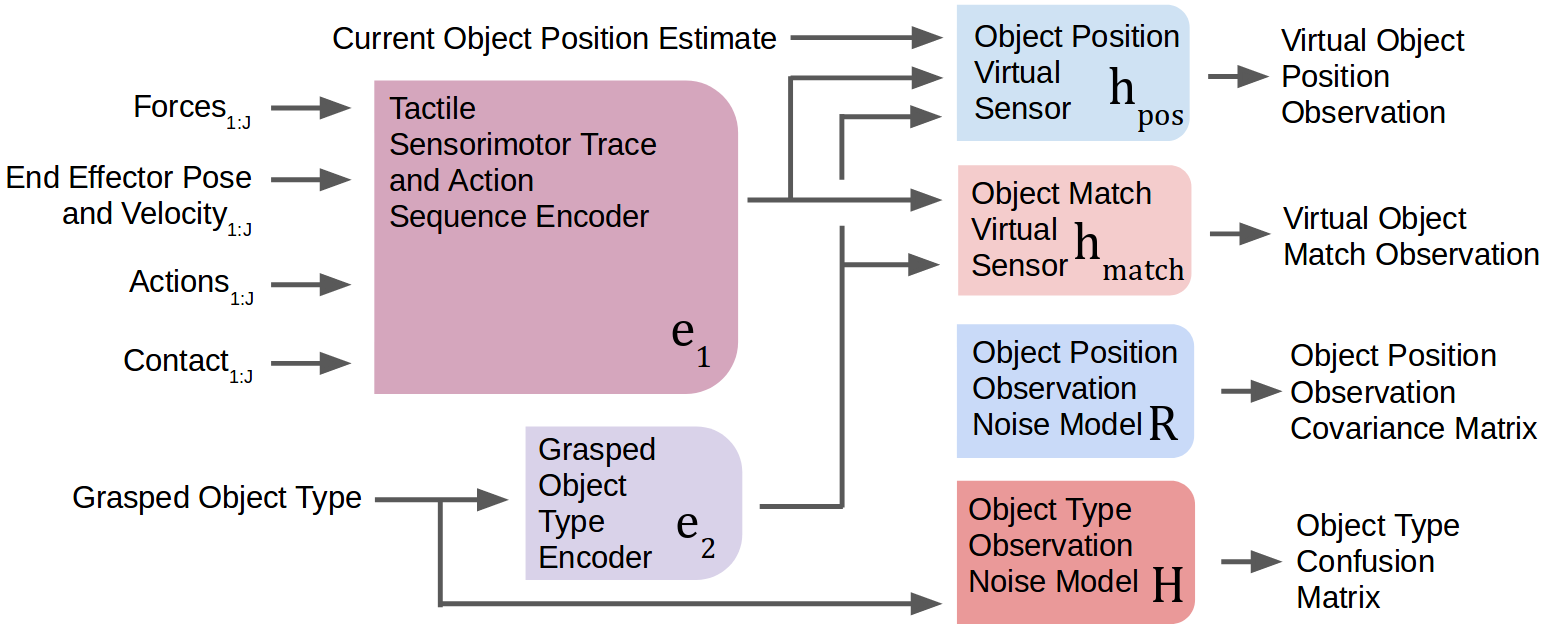}
    \caption{Schematic of the Architecture of the Position and Type Filters}
    \label{fig:arch}
\end{figure}

As described in Section~\ref{subsection:task}, the differentiable Kalman filter requires a virtual sensor $h_{pos}()$ and the position noise model $\mathbf{R}$. The differentiable histogram filter requires a virtual sensor $h_{match}()$ and the type noise model in $H(o_{match}|c^i, c^{peg})$. The components of the differentiable Kalman and histogram filters are illustrated in Fig.~\ref{fig:arch}.

As shown in the figure, both $h_{pos}()$ and $h_{match}()$ share a common set of encoders $e_1()$, and $e_2()$. There are two sub-modules in $e_1()$. The first is a force encoder, which is a $5$-layer $1$-d convolutional neural network, mapping each force measurement in the sensorimotor trace $\mathbf{z}_{1:J}$ to a vector. The second sub-module is a $4$-layer transformer decoder network \cite{transformer} with a $\max$ operation along the time dimension of its output sequence. The output of the force encoder is fed into the transformer network sub-module along with $\mathbf{u}_{1:J}$ and the rest of the measurements in $\mathbf{z}_{1:J}$. Encoder $e_2()$ is an embedding layer used to encode $c^{peg}$. The covariance matrix $\mathbf{R}$ and the confusion matrices in $H()$ are also embedding layers. The virtual sensors $h_{pos}()$ and $h_{match}()$ are $3$-layer fully-connected networks with skip connections between the first $2$ layers. For our experiments we used a $64$-dimension encoding for the outputs of $e_1()$ and $e_2()$.

To train the differentiable Kalman and histogram filters, we collected a dataset of interaction trajectories with all possible pegs and holes using randomly sampled robot end-effector commands (see Section\ref{section:filter_training}). Each data point contains the following information: 1) a tactile sensorimotor trace $\mathbf{z}_{1:J}$, 2) an action sequence $\mathbf{u}_{1:J}$, 3) the type of the robot's peg $c^{peg}$, 4) an approximate initial belief of hole position $\{\bm{\mu}_0, \mathbf{\Sigma}_0\}$,
5) an approximate initial belief about hole type $\bm{\xi}_{0}$,
and 6) the hole type $c$ and position $\mathbf{p}$. As detailed in Section~\ref{section:filter_training}, $\{\bm{\mu}_0, \mathbf{\Sigma}_0\}$ and $\bm{\xi}_{0}$ are training hyperparameters. These inputs are passed to the differentiable Kalman filter as described in Algorithm 2 and to the differentiable histogram filter as described in Algorithm 3, resulting in an updated belief of hole position $\{\bm{\mu}_{1},\mathbf{\Sigma}_{1}\}$ and hole type $\bm{\xi}_{1}$.

To train the differentiable filters, we minimize the negative log-likelihood of the true hole state with the updated belief:
\begin{equation} \label{eq:filter_loss}
\begin{split}
    Loss_{filters} =& \frac{1}{2} \ln |\mathbf{\Sigma}_{1}| + \frac{1}{2} (\mathbf{p} - \bm{\mu}_{1})^T \mathbf{\Sigma}_{1}^{-1} (\mathbf{p} - \bm{\mu}_{1})\\
    & - \sum_{c^{\prime} \in C} \mathbb{I}_{c^{\prime} = c} \ln(\xi_{1,c^{\prime}})\\
    &-\mathbb{I}_{c = c^{peg}} \ln(p(c = c^{peg}))\\
    &- (1 - \mathbb{I}_{c = c^{peg}}) \ln(p(c\neq c^{peg})).
\end{split}
\end{equation}
The first two terms train $h_{pos}()$ and $\mathbf{R}$. The third term is to train the observation model $H()$. The last two terms define the cross entropy loss used to train the virtual sensor $h_{match}()$, where $p(c\neq c^{peg})$ and $p(c = c^{peg})$ are calculated by performing a softmax operation on the outputs of $h_{match}()$.

\section{Experimental Setup and Implementations}


\subsection{Experimental Setup}
We propose a hierarchical approach with high and low level policies to solve assembly problems. To evaluate the method, we simulated a robotic peg insertion task with $3$ different peg-hole types and up to $5$ pegs and $5$ hole boxes in the environment, as illustrated in Fig.~\ref{fig:teaser}.
The clearance of the pegs with their corresponding holes was $1$mm. The simulation was implemented using the Mujoco simulator~\cite{mujoco} and Robosuite environments~\cite{robosuite}. We used a $7$-DOF Franka-Emika Panda robotic arm for the task. To transform the robot's actions $\mathbf{u}_{1:J}$ from the low level policy into desired torque commands, we use operational space control~\cite{opspace} similar to~\cite{vices} with an update frequency of 10Hz; the torque commands are sent to the robot at $500$Hz. We simulated a noisy hole detector with a detection error up to $2$cm. Specifically, the detected hole positions are randomly sampled \textcolor{black}{from $U(-\mathbf{2}cm,\mathbf{2}cm) + \mathbf{p}$, where $\mathbf{p}$ is the ground truth position of the hole}. The detected positions are used as the robot's initial estimate of hole positions, e.g., $\bm{\mu}_{0}^i$ for the $i$-th hole. For the robot's initial belief about the $i$-th hole type, we used an uninformative prior $\xi_{0,c}^i = \frac{1}{C}$, $\forall c \in \{1, \cdots, C\}$. \textcolor{black}{To simulate a noisy force-torque sensor, we added samples to the measurements from a zero-mean Gaussian with a variance according to the characteristics of a Robotiq FT 300-S \cite{robotiq}.}

\subsection{Low Level Policy}
The overall approach proposed in this paper is agnostic to the choice of the low level policy $\pi_{low}$. To keep our method simple, we use an open-loop policy $\pi_{low}$ for all experiments, which is similar to~\cite{spiral_search} and described in Algorithm 4. $\pi_{low}$ performs a spiral search around the current hole position estimate with random wiggling to prevent the robot from getting stuck. 
In Algorithm 4, $r_{max}$ is the maximum radius of the spiral search, $n_{rot}$ is the desired number of spiral revolutions during execution, $\delta z$ is a constant position error which keeps the robot pressing against the peg hole box, and $\sigma$ is a parameter which determines the scale of wiggling. For our experiments, we used the values: $r_{max} = 1.5$cm, $n_{rot} = 2$, $\delta z=0.4$cm, $\sigma = 0.125$cm. In our experiments, the low level task horizon $J$ was $100$, corresponding to a $10$ second interaction with the object. We performed $100$ low level policy rollouts on the above described peg insertion tasks where the grasped peg always matched the hole, and measured the success rate as $\alpha=0.34$.

\begin{algorithm}[t]
\setstretch{1.15}
\LinesNumbered
\SetAlgoLined
\KwData{$\mathbf{x}_j, \mathbf{p}_{t}^i, j, J$}
\KwResult{$\mathbf{u}_j$}
    $u_{spiral|x} = \frac{j r_{max}}{J}cos(\frac{2 \pi j  n_{rot}}{J})$\\
    $u_{spiral|y} = \frac{j r_{max}}{J}sin(\frac{2 \pi j  n_{rot}}{J})$\\
    $u_{spiral|z} = -\delta z$\\
    $\mathbf{u}_{wiggle} \sim N(\mathbf{0}, \sigma^2 \mathbf{1})$\\
    $u_{wiggle|z} = |u_{wiggle|z}|$\\
    $\mathbf{u}_j = \mathbf{u}_{spiral} + \mathbf{u}_{wiggle} + \mathbf{p}_{t}^i - \mathbf{x}_j$\\
\caption{Spiral search policy $\pi_{low}$}
\end{algorithm}


\subsection{Differentiable Filter Training}
\label{section:filter_training}
To train the differentiable Kalman and histogram filters, we collected a dataset of $3000$ interactions of peg fitting. Each interaction lasted $20$ seconds ($J=200$) or less if the robot achieved insertion before that time. To avoid data imbalance, half of the $3000$ interactions were between matched peg and hole shapes, and the other half were between mismatched shapes. During each interaction, the robot initially moved to a position sampled from the simulated hole detector, and then performed a sequence of randomly generated actions, similar to Line $4$ in Algorithm 4.

We trained the differentiable filters using the Adam optimization algorithm \cite{adam} with a learning rate of $0.0001$ for $5000$ epochs. During training, data points are generated by sampling sub-sequences of length $l$ from the collected interactions where $l$ was a randomly selected integer in the range $[20, 100]$. To generate $\{\bm{\mu}_0, \mathbf{\Sigma}_0\}$, we used $\bm{\mu}_{0} \sim \mathcal{U}(-\mathbf{2}cm
, \mathbf{2}cm)$\textcolor{black}{$+ \mathbf{p}$} and an isotropic covariance matrix $\mathbf{\Sigma}_0 = \sigma_{init}\mathbf{I}$ where $\sigma_{init} = 1cm^2$. We used this same covariance matrix as the initial uncertainty estimate for our simulated vision based hole detector at test time. To generate $\bm{\xi}_{0}$, we sampled three numbers (for the three possible hole shapes) from $\mathcal{U}(0,1)$, and applied $L_0$ normalization.

\section{Results}

We investigate $3$ questions about how our hierarchical approach performs in object fitting.
\begin{enumerate}
    \item Does probabilistic {\em sequential\/} estimation of hole positions outperform frame-by-frame approaches?
    \item If the above is true, does the improvement in the hole position estimate translate to increased performance of the low level task execution?
    \item Does the use of a tactile sensorimotor trace help in hole type estimation, i.e., help to distinguish a high level task failure (wrong hole match) from a low level task failure (failed peg-hole fitting), thus reducing the number of fitting attempts?
\end{enumerate}

\subsection{Sequential Estimation of Object Positions}
\label{section:exp_sequential}
Our hypothesis is that sensorimotor traces collected during physical interaction with a hole provide valuable information for updating the robot's belief about the hole state. In this experiment, we focus on how well the hole position can be estimated from a sequence of high level task steps. We compare the proposed approach with a frame-by-frame baseline that ignores the prior belief, instead using the output of a virtual position sensor as the current position estimate.

\begin{figure}
    \centering
    \includegraphics[width=\columnwidth]{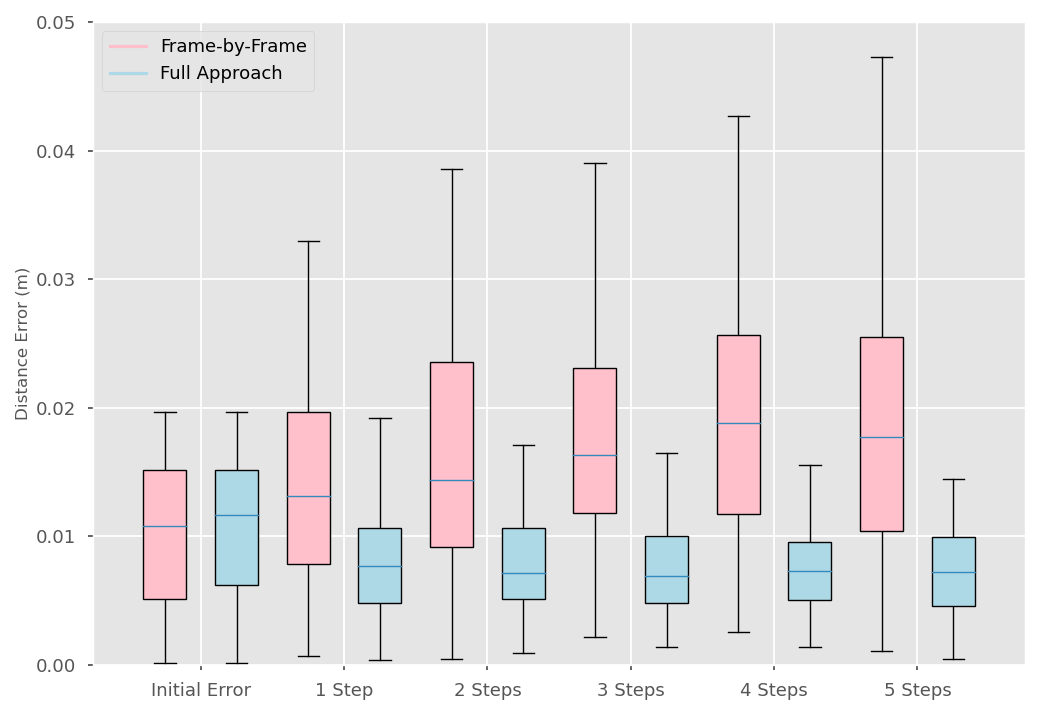}
    \caption{We compared our approach (Full Approach) on the metric of accurate hole position estimation where the robot takes $5$ high level task steps to improve its estimate. The $y$-axis indicates the position estimate error in meters and the $x$-axis indicates the number of high level steps performed. 
    }
    \label{fig:belief}
\end{figure}

We train a virtual sensor $h_{pos|baseline}()$ using the same architecture as $h_{pos}()$ but estimate hole positions relative to the robot's final end-effector position $\mathbf{x}_J$. The current estimate of hole position $\hat{\bm{p}}_t$ equals $h_{pos|baseline}() + \mathbf{x}_J$. We compared both approaches on a position estimation task where the robot can make five fitting attempts to interact with a hole (which corresponds to five high-level steps). Fig.~\ref{fig:belief} summarizes the performance of both approaches on this task averaged over $100$ trials. The results show that with the proposed sequential approach, the position estimation error quickly decreases to around \textcolor{black}{$0.7$}cm on average from $1.1$cm initially. Meanwhile, the estimation error with the frame-by-frame approach fluctuates around a large error with a higher estimation variance than the sequential approach.

This result demonstrates that the hole position estimate improves when updated from haptic sensor traces during contact-rich interactions, even though the robot failed at the low level task. Since $\pi_{low}$ depends on this estimate, we expect that the number of necessary high level task steps decreases when comparing our proposed method against the baselines.
%
%
To test this hypothesis, we evaluated our approach's performance on a peg insertion task where the attempted hole always matched the robot's peg. 

\begin{figure}
    \centering
    \includegraphics[width=\columnwidth]{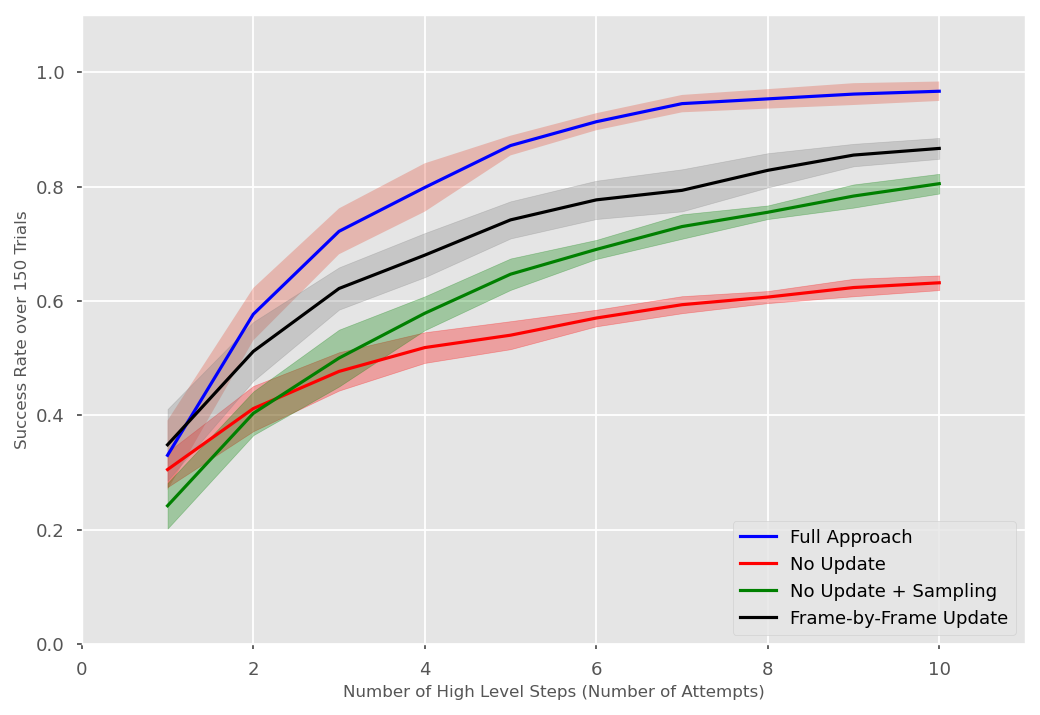}
    \caption{We compared our approach (Full Approach) to a set of baselines on an object fitting task where the robot interacted with a hole which fits its peg and takes up to $10$ high level steps. The $y$-axis measures the success rate over the $150$ trials and $x$-axis indicates the number of high level task steps performed.
    }
    \label{fig:post_vs_prior}
\end{figure}

We implemented three baselines for this experiment: 1) the robot starts each low level trial at the initial position estimate $\hat{\bm{p}}_t = \bm{\mu}_0$; 2) the robot starts at a position sampled from the robot's initial belief distribution $\hat{\bm{p}}_t \sim \mathcal{N}(\bm{\mu}_0, \mathbf{\Sigma}_0)$; 3) position is estimated using the frame-by-frame approach described in Section~\ref{section:exp_sequential}. Fig.~\ref{fig:post_vs_prior} shows the success rate of each approach over $150$ trials with $5$ different random seeds for introducing noise in the low-level policy (Line 4, Algorithm 4). The results suggest that without updating the hole position belief, the robot can complete the task up to $80\%$ of the time using sampling. But with our full approach, the robot almost always completes the task within $10$ high level steps. A frame-by-frame approach also improves the success rate but performs significantly worse than the full approach, due to its high variance \textcolor{black}{in position estimation}.
These experiments demonstrate the value of sequentially updating the belief about object positions using sensorimotor traces collected from failed low level task executions.


\subsection{Object Fitting with Uncertainty}
Our proposed approach handles uncertainty not only about hole positions but also hole types. Using sensorimotor traces collected during high level task steps, we also update the belief about hole types. This will inform the high-level policy for potentially choosing a different object to interact with next. Our hypothesis is that the proposed model requires fewer fitting attempts to complete an entire assembly task.
To test this hypothesis, we design the fitting task as follows. There are $5$ holes in the environment of $3$ possible hole types, and the robot has $5$ pegs known to match the $5$ holes in the environment. The robot randomly grasps a new peg from this set each time it achieves fit with the previously grasped peg. This is repeated until fit has been achieved with all $5$ holes.

We implemented two baselines. In the first baseline \texttt{Failure only}, the robot does not use the sensorimotor trace for belief updates. It maintains a constant belief about hole positions as initialized from the vision sensor. However, it updates its belief about hole types based on the outcomes of fitting attempts using the transition model in Eq.~\ref{eq:dynamics_model}. For the second baseline \texttt{Failure + Position}, the robot updates the belief about hole types in the same way as \texttt{Failure only}, but uses the sensorimotor trace to update the hole position belief. \textcolor{black}{We refer to our proposed method as \texttt{Full Approach} in which the sensorimotor trace is used to update the belief about both, hole type and position.}

\begin{figure}
    \centering
    \includegraphics[width=\columnwidth]{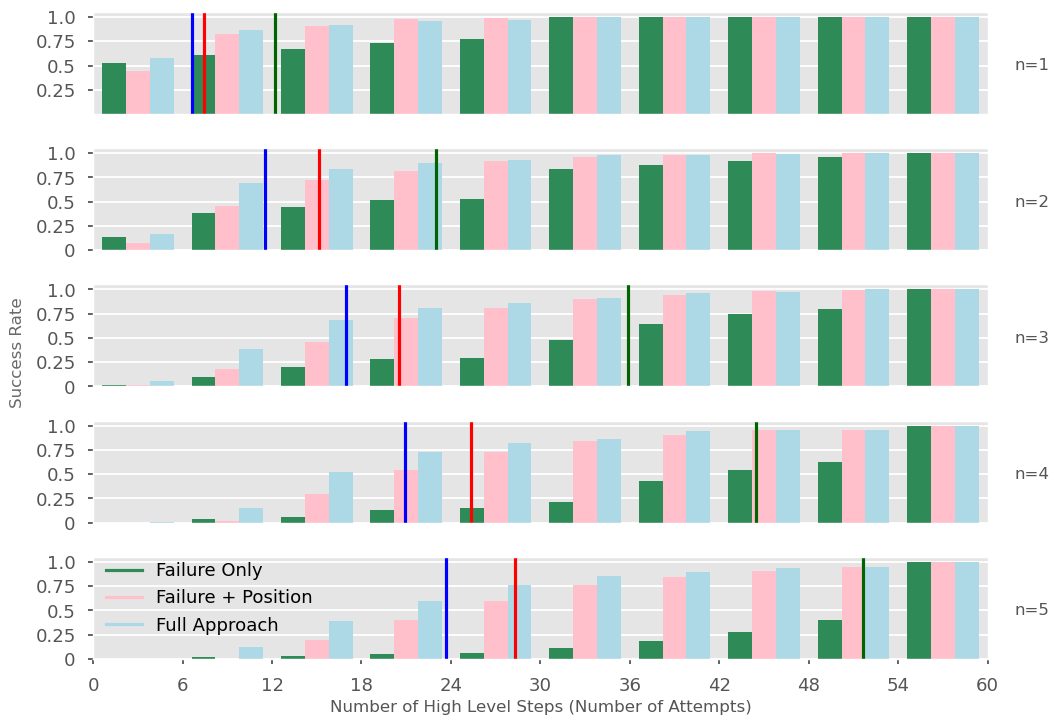}
    \caption{We compared our approach (Full Approach) with two baselines on an object fitting tasks with $5$ holes of $3$ possible hole types. Each row corresponds to the success rate of inserting the first $n$ pegs over $100$ trials plotted over the number of fitting attempts on the x-axis. The histogram groups the number of attempts into bins on the $x$-axis with a bin size of $6$. The overlaid vertical lines indicate the mean number of attempts required by each approach to insert the first $n$ pegs. From the results, we can see that our approach needs on average many fewer attempts to fit all 5 pegs than the baselines.
    }
    \label{fig:trace}
\end{figure}

Fig.~\ref{fig:trace} shows the cumulative number of high level steps it took to insert the first $n$ pegs for all three approaches over $100$ trials. During experiments, we capped the number of high level task steps for each object fitting task to $30$, upon which a human intervenes and \textcolor{black}{inserts the current peg in an empty peg hole which matches the robot's current peg. Then the robot continues on to the next peg}. \texttt{Failure Only} required human intervention \textcolor{black}{in $26.4\%$ of the 500 peg insertion tasks performed in the 100 trials,} \texttt{Failure + Position} required intervention in \textcolor{black}{$1.0\%$} of the tasks; and \texttt{Full Approach} required intervention in \textcolor{black}{$2.2\%$} of the tasks. The inferior performance of \texttt{Failure Only} indicates that sensorimotor signals are critical to our approach for dealing with uncertainty in hole position and for recovering from failure. \textcolor{black}{The marginally higher intervention rate of the \texttt{Full Approach} indicates that in rare cases our proposed approach converges on the wrong object type, but using this approach comes with the benefit that on average it accomplishes the task faster than the \texttt{Failure + Position} baseline. This} suggests that the sensorimotor signals are informative not only for hole positions, but also for hole types. We expect this to become even more significant, the higher the number of peg-hole pairs.

\section{Conclusion}
We propose a method for accomplishing robotic object fitting tasks, where the robot does not know with certainty the types and positions of objects in the environment. We propose a hierarchical approach where the high level task has to find matching objects and the low level task has to fit them. Specifically, we propose to use tactile sensorimotor traces from fitting attempts to update the belief about the state of the environment. In experiments, we evaluated the effectiveness of our approach at leveraging tactile sensorimotor traces from failed fitting attempts to improve the robot's performance over time. Results showed our approach achieves higher precision in estimating object position, and accomplishes object fitting tasks faster than baselines. While our experiments demonstrate promising results in simulated environments with simulated noise, we have yet to demonstrate this approach on a real robot which we look forward to post-COVID. For future work, we will explore more complex assembly tasks, such as bimanual manipulation of assembly objects, for which in-hand object pose uncertainty also needs to be handled.

\bibliographystyle{IEEEtranN}
\bibliography{sensorimotor_search}

\end{document}